\documentclass[letterpaper]{article} 
\usepackage{aaai25}  
\usepackage{times}  
\usepackage{helvet}  
\usepackage{courier}  
\usepackage[hyphens]{url}  
\usepackage{graphicx} 
\usepackage{natbib}  
\usepackage{caption} 
\usepackage{algorithm}
\usepackage{algorithmic}

\usepackage{booktabs}
\usepackage{multicol}
\usepackage{multirow}
\usepackage{pifont}
\usepackage{amssymb}
\usepackage{amsmath}
\usepackage{cleveref}
\usepackage{subcaption}
\usepackage{xcolor}
\usepackage{bm}
\usepackage{adjustbox}
\usepackage{tabularx}

\def\mytracker{TemTrack}
\def\mymodule{Temporal Module}
\def\mytoken{track token}
\usepackage{newfloat}
\usepackage{listings}
\DeclareCaptionStyle{ruled}{labelfont=normalfont,labelsep=colon,strut=off} 
\floatstyle{ruled}
\newfloat{listing}{tb}{lst}{}
\floatname{listing}{Listing}
\title{Robust Tracking via Mamba-based Context-aware Token Learning}
\author{
    Jinxia Xie\textsuperscript{\rm 1,2}, Bineng Zhong\textsuperscript{\rm 1,2}\thanks{Corresponding Author}, Qihua Liang\textsuperscript{\rm 1,2}, Ning Li\textsuperscript{\rm 1,2}, Zhiyi Mo\textsuperscript{\rm 3}, Shuxiang Song\textsuperscript{\rm 1,2}
}
\affiliations{
    \textsuperscript{1}Key Laboratory of Education Blockchain and Intelligent Technology Ministry of Education, \\
    Guangxi Normal University, Guilin 541004, China\\
    \textsuperscript{2}Guangxi Key Lab of Multi-Source Information Mining \& Security, Guangxi Normal University, Guilin 541004, China\\
\textsuperscript{3}Guangxi Key Laboratory of Machine Vision and Intelligent Control, Wuzhou University, Wuzhou 543002, China\\
{\tt\small xie\_jx@stu.gxnu.edu.cn, bnzhong@gxnu.edu.cn, qhliang@gxnu.edu.cn, ningli65536@mailbox.gxnu.edu.cn, zhiyim@gxuwz.edu.cn, songshuxiang@mailbox.gxnu.edu.cn}
}

\usepackage{bibentry}

\begin{document}

\maketitle

\begin{abstract}
How to make a good trade-off between performance and computational cost is crucial for a tracker. However, current famous methods typically focus on complicated and time-consuming learning that combining temporal and appearance information by input more and more images (or features). Consequently, these methods not only increase the model's computational source and learning burden but also introduce much useless and potentially interfering information. To alleviate the above issues, we propose a simple yet robust tracker that separates temporal information learning from appearance modeling and extracts temporal relations from a set of representative tokens rather than several images (or features). Specifically, we introduce one {\mytoken} for each frame to collect the target's appearance information in the backbone. Then, we design a mamba-based {\mymodule} for {\mytoken}s to be aware of context by interacting with other {\mytoken}s within a sliding window. This module consists of a mamba layer with autoregressive characteristic and a cross-attention layer with strong global perception ability, ensuring sufficient interaction for {\mytoken}s to perceive the appearance changes and movement trends of the target. Finally, {\mytoken}s serve as a guidance to adjust the appearance feature for the final prediction in the head. Experiments show our method is effective and achieves competitive performance on multiple benchmarks at a real-time speed. Code and trained models will be available at 
\url{https://github.com/GXNU-ZhongLab/TemTrack} .

\end{abstract}

\section{Introduction}
\begin{figure}[t] 
    \centering
    \includegraphics[height=5.2cm]{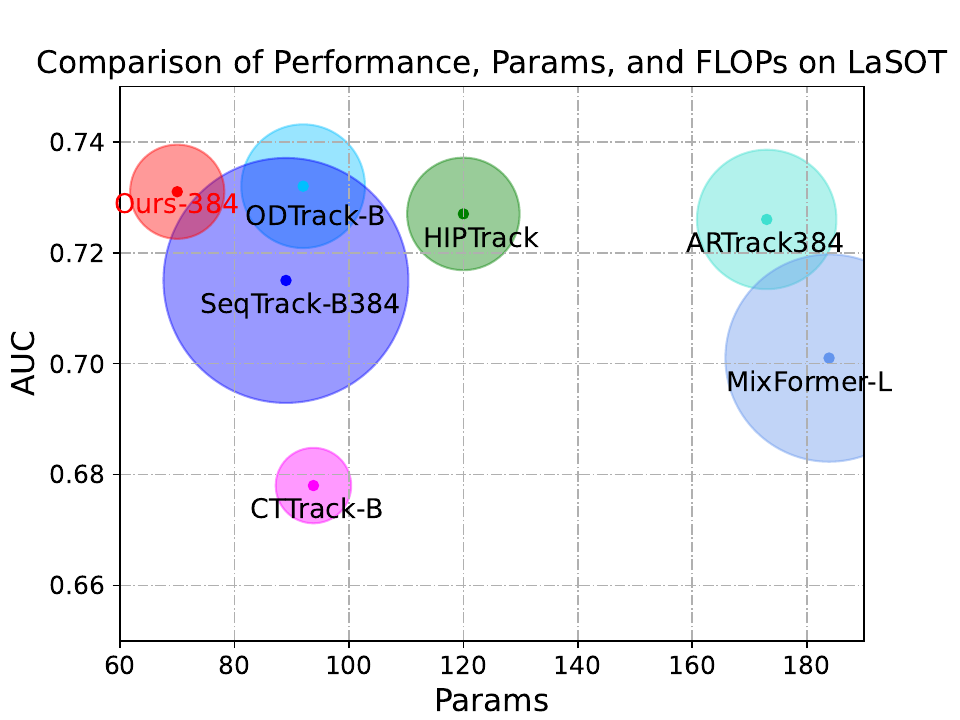}
    \caption{
    Comparison of AUC, Params, and FlOPs of recent SOTA trackers.
    The trackers that lower Params and higher AUC is closer to top-left corner.
    The size of a circles represents the tracker’s FLOPs.
    }
    \label{fig1}
\end{figure}
    Visual tracking is one of the fundamental tasks in computer vision, widely used in many fields, such as mobile robotics\cite{mobile_robotics}, video surveillance\cite{surveillance0, surveillance1}, and autonomous driving\cite{autodriving0}.
    However, there are many challenges during the tracking process that affect the robustness of the trackers, such as occlusion, drastic appearance changes, and deformation.

    Therefore, many methods\cite{SiamRPN++, SiamFC++, Siamban, sparseTT, CTTrack, AQATrack, hu2024} are proposed and attempt to overcome the above challenges.
    These methods can be roughly divided into two types: trackers focused more on appearance and trackers combined appearance with temporal information. 
    For the first type of trackers\cite{ostrack, transt}, they focus on building a more robust appearance model via a stronger backbone or a more efficient feature fusion method for template and search image.
    However, it's difficult for this type of trackers\cite{SiamFC,ostrack, simtrack, GRM} to recognize the correct target when facing severe appearance changes or interference from similar objects.
    Recently, the visual tracking community pay more attention to extract temporal context to mitigate the above difficulty.
    Many second type of trackers\cite{odtrack, artrackv2, VideoTrack1, hiptrack,mixformer,MixViT} arise, combining appearance and temporal information.
    Thanks to introducing temporal information, these trackers perceive the appearance changes and motion trends of the target, and achieve competitive performance.
    However, they usually focus on complicated and time-consuming learning, inputting more images (or features), and leading the model more cumbersome and clumsy.
    Specifically, they\cite{stark, seqtrack, SwinTrack} need to select additional images besides one template and one search image, which requires controlling thresholds or manually crafting components for the selection strategy.
    These processes are tedious and not flexible.
    Furthermore, even when simple methods are employed for selecting images (or features), the large input size can significantly increase the computational resource and learning burden, and lead to heavy training costs.
    For instance, SeqTrack\cite{seqtrack} inputs two templates with the same size as the search image, and its number of floating point operations (FLOPs) is 148G, which is nearly three times our tracker (55.7G), as shown in \cref{tab:fps}.
    And ODTrack\cite{odtrack} input three templates and one search image, which is time-consuming for model learning.
    Finally, increasing the number of images may introduce more useless or potentially interfering information leading to a suboptimal tracking result.
    The comparison with recent context-aware trackers of params, FLOPs, and performance on LaSOT\cite{lasot} is shown in \cref{fig1}.

    To make a good trade-off between performance and computational cost, we propose a simple and efficient context-aware tracker, named {\mytracker}, which separates temporal information learning from appearance modeling and learns contextual information from a set of {\mytoken}s instead of images.
    In this way, it can alleviate the computational source and learning burden caused by inputting too many images, and the backbone network can focus more on learning the target appearance and modeling the relationship between templates and search images.
    Specifically, we introduce a {\mytoken} for each frame and feed it into the backbone alongside template and search tokens.
    Each {\mytoken} is responsible for collecting the appearance information of the target.
    After the backbone, each token contains the appearance information of the target in that frame.
    Then, we set a sliding window with a size of $m$. 
    The {\mytoken}s in the sliding window are fed into a mamba-based {\mymodule} for temporal context learning.
    This module consists of a mamba layer with a autoregressive characteristic and a cross-attention layer with strong global perception, which ensures sufficient interaction for {\mytoken}s to perceive the appearance changes and movement trends of the target.
    After interaction with the other tokens, the {\mytoken} contains temporal information.
    Finally, we use the {\mytoken} to adjust search features through simple operations, and then search features are fed into the head to predict the target's position and size.
    To summarize, the main contributions of this work are as follows: 
    \begin{itemize}
        \item To make a good trade-off between performance and computational cost, we propose a simple but robust tracker, which separates temporal information learning from appearance modeling, extracting temporal relations from a set of representative tokens in a sliding window fashion.
        \item We develop an efficient mamba-based module for modeling contextual information, named {\mymodule}. This module consists of mamba and attention mechanism, combining long sequence modeling and global perception capabilities. 
        \item We conduct detailed experiments to verify the effectiveness of our temporal context information modeling method.
        The results demonstrate our method achieve a new state-of-art on multiple benchmarks. 
    \end{itemize}
\begin{figure*} 
    \centering
    \includegraphics[width=16cm]{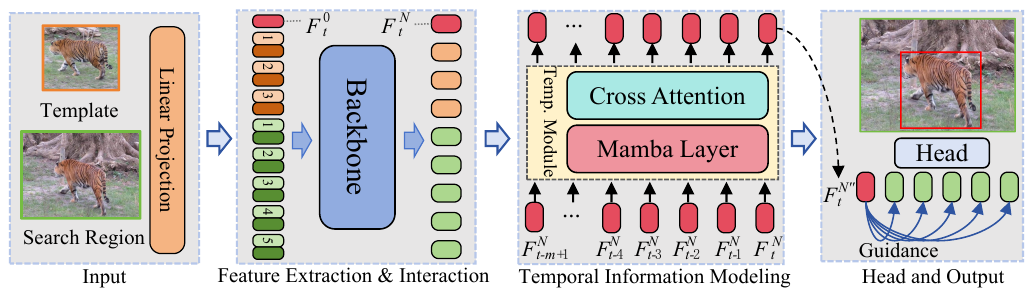}
    \caption{
    Overview of the proposed tracker {\mytracker}. 
    The tracker's workflow is depicted from left to right, including feature extraction \& interaction, temporal information modeling, and the final head stage.
    First, we add a {\mytoken} $\bm F_t$ concatenating with template and search tokens to gather the target's appearance in the backbone.
    Furthermore, we develop a {\mymodule} to associate {\mytoken}s to dig temporal information.
    Finally, the {\mytoken}s guide the adjustment of the search features to achieve more accurate predictions in the head network.
    }
    \label{fig2}
\end{figure*}
\section{Related Work}
\textbf{Trackers Focusing on Appearance Modeling.}
With the development of deep learning and the introduction of attention mechanisms, significant progress has been made in visual object tracking.
Many trackers\cite{ostrack, cswintt, huxiantao} focus more on appearance modeling, use a powerful backbone, and design a more effective module for feature confusion.
SiamFC\cite{SiamFC} based AlexNet\cite{alexnet} design a Siamese network to extract features and use fully-convolutional deep networks to fuse the feature.
TransT\cite{transt} uses ResNet\cite{resnet1} as the backbone and introduces the attention to design a correlation module.
Thanks to Swin Transformer\cite{swintransformer} as the backbone, SwinTrack\cite{SwinTrack} achieves outstanding performance.
One of the most successful trackers is OSTrack\cite{ostrack}, which uses ViT\cite{vit} as the backbone and proposes a simple yet effective one-stream tracking paradigm.
Thus some trackers\cite{simtrack, evptrack} adopt the one-stream paradigm and introduce other strong transformer variants as backbone, making significant progress.
Our tracker adopts the great one-stream paradigm to model the appearance.

\textbf{Trackers Combining Appearance and Temporal Context.} 
Temporal information captures the appearance changes and motion patterns of targets, playing a crucial role in enhancing robustness against drastic appearance changes and interference from similar objects.
So many trackers\cite{stark, aiatrack, xue2024, artrackv2} combine appearance and temporal information to help trackers achieve more accurate tracking.
Most trackers introduce the temporal information by updating a dynamic template image, which requires controlling thresholds or manually crafting components, such as MixFormer\cite{mixformer}, CTTtrack\cite{CTTrack}, and SeqTrack\cite{seqtrack}.
In addition, UpdateNet\cite{updateNet} estimates an optimal template from several images for the next frame.
STMTrack\cite{STMTrack} uses a memory network to integrate historical features.
VideoTrack\cite{VideoTrack1} mining temporal information from video clips.
Some trackers\cite{TCTrack++, evptrack, AQATrack,odtrack} transmit temporal context to enhance the tracker's ability to distinguish targets.
Although the above trackers achieve good performance, these models are usually more complex due to the need to design strategies to select images and withstand more learning and computational burden from inputting more images (or features).
So we design a simple yet robust tracker with less computational cost, without updating strategy or inputting more images.

\textbf{Mamba in Visual Task.}
Recently, the mamba with autoregressive characteristic become famous for its linear complexity and is introduced into many visual tasks.
In upstream tasks, Vmamba\cite{vmamba} constructs a hierarchical vision model based on mamba with a four-direction scanning strategy.
Vision Mamba\cite{vim} proposes a bidirectional state space model referred to ViT's\cite{vit} pipeline.
LocalMamba\cite{localmamba} incorporates local inductive biases to enhance visual mamba models.
In medical object segmentation, numerous studies adopt mamba-based models, such as U-Mamba\cite{u-mamba} and SegMamba\cite{segmamba}.
So many success models demonstrate mamba's outstanding long-sequence processing capabilities.
In this work, we integrate mamba into {\mymodule} to ensure sufficient interaction between {\mytoken}s.

\section{Our method}
    This section offers a concise and lucid description of the proposed robust \underline{tem}poral tracker, called \mytracker.
    First, we describe the tracking framework of {\mytracker}.
    Then, we introduce the main components, including a backbone and the {\mymodule}.
    Finally, we briefly describe the guidance from {\mytoken} to appearance, head, and loss function.

\subsection{Overview.}
    The framework of the {\mytracker} is demonstrated in \cref{fig2}, whose main components are a strong backbone, a mamba-based {\mymodule}, and a head.
    The input for the tracker is a pair of images, namely one template image $\mathbf{Z}\in\mathbb{R}^{h_{z}\times W_{z}\times3}$ and one search image $\mathbf{X}\in\mathbb{R}^{h_{x}\times W_{x}\times3}$.
    These two images are embedded and then concatenate with a {\mytoken} to be fed into the backbone.
    The {\mytoken} is one of the key components of {\mytracker}, whose responsibility is to gather the target's appearance from the image tokens in the backbone and learn the temporal context in the {\mymodule}.
    Before the head, the appearance (search features) is adjusted by {\mytoken} with temporal information, and fed into the head for the final prediction.
    
    \subsection{Feature Extraction and Relation Modeling.}
    OStrack\cite{ostrack} proves that joint feature extraction and relationship modeling can enable sufficient interaction between templates and search features.
    Trackers\cite{GRM} modeled using this approach can greatly improve their ability to discriminate targets.
    They usually use Vanilla ViT\cite{vit} as a backbone to complete the above goals.
    ViT embeds the images to patches with size 16 × 16 at once, which loses a lot of information about adjacent patches\cite{AQATrack}.
    To avoid this issue, we choose Fast-iTPN\cite{fastitpn} as the backbone, which performs downsampling twice via two merge layers before global attention.
    After downsampling, the features shape of the template and search are $\bm{F}_{z}^0\in\mathbb{R}^{N_{z}\times D}$ and $\bm{F}_{x}^0\in\mathbb{R}^{N_{x}\times D}$, respectively.
    Here, $N_{z}=h_{z}W_{z}/16^{2}$, $N_{x}=h_{x}W_{x}/16^{2}$, $D=512$.
    So the patch size after downsampling is the same as other trackers, both are 16 × 16.
    To learn temporal information with a small cost in {\mymodule}, and also focus more on modeling the target appearance and relation between the template and search, we introduce one {\mytoken} $\bm{F}_{t}^0\in\mathbb{R}^{1\times D}$ for each pair of images, where $t$ means at $t$ frame.
    The remaining operation in the backbone can be summarized as the following formula:
    \begin{equation}
        \begin{split}
            &\bm{F}_{tzx}^0 = Concat(\bm{F}_{t}^0 , \bm{F}_{z}^0, \bm{F}_{x}^0), \\
            &\bm{F}_{tzx}^n = Backbone(\bm{F}_{tzx}^{n-1}), n=1...N, 
            \label{eq:Backbone}
        \end{split}
    \end{equation}
    where $N$ is the number layer of global attention in the backbone. Refer to Fast-iTPN\cite{fastitpn} for more details.
    
\subsection{Temporal Information Learning.}
\begin{figure}[t] 
    \centering
    \includegraphics[width=6cm, height=6cm]{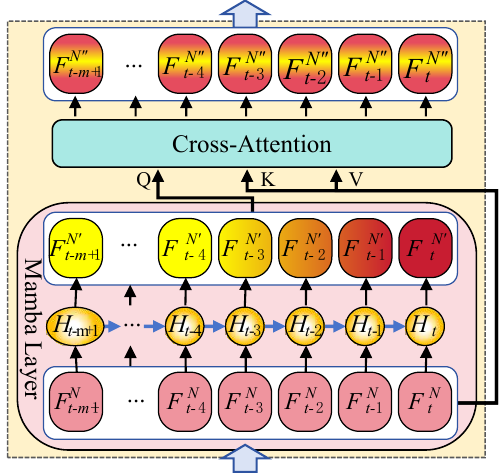}
    \caption{The schematic diagram of the Mamba\_Cross. The $\bm{F}_t^N$ and $\bm{H}_t$ indicate {\mytoken} and the hidden state at the $t$ frame. After this module, the {\mytoken} $\bm{F}_t^{N''}$ gathers the appearance of the previous frames within a sliding window.}
    \label{fig3}
\end{figure}
    To demonstrate the superiority of our method, we develop \textbf{three variants} of the {\mymodule}.
    Each variant is composed of two layers, i.e., \textit{Mamba\_Cross}, \textit{Self\_Cross}, and \textit{Self\_Self}.
    All of them outperform most trackers, the results are shown in \cref{tab:temporal_module}.
    The outstanding performance of the above variants demonstrates that our method can effectively associate contextual information through {\mytoken}s. 
    
    The input of the {\mymodule} is a set of historical {\mytoken} $\mathbf{T}$ containing the appearance information of the target at various times:
    \begin{equation}
        \mathbf{T} = Concat(\bm{F}_{t-m+1}^N, ... , \bm{F}_{t-1}^N, \bm{F}_{t}^N),
    \end{equation}
    where $m$ is the size of the sliding window. 
    In {\mymodule}, the {\mytoken} $\bm F_t$ interacts with other {\mytoken}s within a sliding window. 
\subsubsection{Mamba\_Cross.}
    To better dig the historical target state implied in {\mytoken}, we combined the mamba\cite{mamba} with long-sequences and autoregressive characteristics\cite{mambaout}, which demonstrate outstanding performance in long sequence tasks.
    So we use mamba in {\mymodule}.
    As illuminated in \cref{fig3}, according to the principle and autoregressive characteristics of mamba, the prediction of {\mytoken} $\bm{F}_{t}^{N'}$ depends on the previously hidden state space $\bm{H}_{t-1}$ and the current {\mytoken} $\bm{F}_{t}^N$.
    After mamba, $\mathbf{T^{'}}$ is fed into a cross-attention layer and interacts with original {\mytoken}s $\mathbf{T}$.
    The process in the Mamba\_Cross can be described as:
    \begin{equation}
    \begin{split}
        & \mathbf{T^{'}} = Mamba(\mathbf{T}), \\
        & \mathbf{T^{''}} = Cross\_Attn(\mathbf{Q=T^{'}, K=T, V=T}),
        \label{eq:mamba_cross}
    \end{split}
    \end{equation}
    where $Q$ is the query, $K$ is the key, and $V$ is the value which is the same in following \cref{eq:self_cross} and \cref{eq:self_self}. 
    Ultimately, $\bm{F}^{N''}_t$ gathers the target's historical appearance changes and the motion trend.
    Thanks to mamba's excellent ability to model sequence, in our experiment, the {\mymodule} variant with mamba achieves the best performance among the three variants, i.e., $74.9\%$ of AO in GOT-10k and  $72.0\%$ of AUC on LaSOT\cite{lasot}, as shown in \cref{tab:temporal_module}.
\subsubsection{Self\_Cross.} 
    This variant consists of a self-attention layer and a cross-attention layer.
    The operation in the Self\_Cross can be described as the following equation:
    \begin{equation}
        \begin{split}
            & \bm{T^{'}} = Self\_Attn(\bm{Q=T}, \bm{K=T}, \bm{V=T}), \\
            & \bm{T^{''}} = Cross\_Attn(\bm{Q=T^{'}}, \bm{K=T}, \bm{V=T}),
            \label{eq:self_cross}
        \end{split}
    \end{equation}
\subsubsection{Self\_Self.}
    In addition, we have also developed a variant that fully utilizes self-attention, namely \textit{Self\_Self}, whose operations can be expressed as the following formula:
    \begin{equation}
        \begin{split}
            & \bm{T^{'}} = Self\_Attn(\bm{Q=T}, \bm{K=T}, \bm{V=T}), \\
            & \bm{T^{''}} = Cross\_Attn(\bm{Q=T^{'}}, \bm{K=T^{'}}, \bm{V=T^{'}}).
            \label{eq:self_self}
        \end{split}
    \end{equation}
        
\subsection{Guidance, Head and Loss.}
    \textit{Guidance.}
    After the {\mymodule}, the $T_t$ merging the historical appearance of the target will guide the search feature to adjust.
    Inspired by STARK\cite{stark}, we calculate the similarity $\bm{S}\in\mathbb{R}^{N_{x} \times 1}$  between search spatial features $\bm{F}_{x}^N\in\mathbb{R}^{N_{x} \times D}$ and {\mytoken} $\bm{F}_{t}^{N''}\in\mathbb{R}^{1 \times D}$.
    The higher the score, the greater the likelihood of the target being located.
    Then use an \textit{element-wise product} to enhance the expression of the search feature.
    
    \textit{Head and Loss.}
    Following the popular trackers\cite{ostrack}, we use the center-based head to predict the tracking box, which includes the position and the scale.
    The center-based head includes two branches, namely classification and regression.
    We use focal loss\cite{focalloss} for classification and combine GIoU loss\cite{giou} and $L1$ loss for regression.
    The total loss $\mathcal{L}$ is calculated as \cref{eq:loss}, which $\lambda_{giou}=2$ and $\lambda_{\mathcal{L}1}=5$.
    \begin{equation}
        \mathcal{L}=\mathcal{L}_{cls}+\lambda_{giou}\mathcal{L}_{giou}+\lambda_{\mathcal{L}_1}\mathcal{L}_1.
        \label{eq:loss}
    \end{equation}

\section{Experiments}
In this section, we introduce the implementation details. Then, we compare our {\mytracker} with SOTA methods on multiple benchmarks. Finally, we show the ablation studies to evaluate the efficiency of the proposed methods. Some tracking results and visualizations are provided to understand how {\mytracker} works.

\subsection{Implementation Details.}
Our tracker is implemented in Python 3.8 using PyTorch 1.13.1. The training is on 4 NVIDIA A10 GPUs and the speed evaluation is on a single NVIDIA V100 GPU.
We present two variants of {\mytracker} with different settings:
\begin{itemize}
    \item \textbf{{\mytracker}-256.} The resolution of template image and search region is 128×128 and 256×256 pixels.
    \item \textbf{{\mytracker}-384.} The resolution of template image and search region is 192×192 and 384×384 pixels.
\end{itemize}

The Fast-iTPN\cite{fastitpn} is used as the backbone for feature extraction and fusion, and the checkpoint of Fast-iTPN-B-224 is loaded to initialize the backbone.

\subsubsection{Training.} Following the mainstream trackers, we use four datasets for training, including COCO\cite{coco}, LaSOT\cite{lasot}, TrackingNet\cite{trackingnet}, and GOT-10k\cite{got10k}.
Common data augmentations are used including bright jittering and horizontal flip.
We train {\mytracker} with AdamW optimizer\cite{adamw}. 
The learning rate of the backbone is 4×$10^{-5}$, and the learning rate of other parameters is 4×$10^{-4}$, and the weight decay is $10^{-4}$. 
The above settings are the same as OSTrack\cite{ostrack}.
Following \cite{AQATrack} and \cite{evptrack}, we sample $n$ video clips for each GPU, which contain $m$ images as search images (all of them with the same template). 
So each GPU holds $n*m$ image pairs, i.e., the batch size is $n*m$. We keep the batch size equal to 32. 
For four GPUs, the total batch size is 128.
Obviously, $m$ is the size of the sliding window and the length of temporal information.
In {\mytracker}, $n$ and $m$ are 4 and 8, respectively.
We train the {\mytracker} with 150 epochs and $60k$ image pairs for each epoch.
We decrease the learning rate by the factor of 10 after the $120th$ epoch.
For the GOT-10k benchmark, we train the model with only $40$ epochs and the learning rate decays at 80\% epochs.

\subsubsection{Inference.} During inference, the {\mytoken} gather the temporal information via the {\mymodule} within a sliding window.
After that, the {\mytoken} that contains historical appearance and motion trend conducts the search feature to adjust.
Following the mainstream tracker\cite{transt, ostrack, AQATrack, evptrack}, we utilize the Hamming window to introduce the positional priors.
Also, we present the Params, FlOPs, and speed of {\mytracker} in the \cref{tab:fps}.
Our {\mytracker}-384 with very less FLOPs runs in real-time at 36 $fps$, faster twice than SeqTrack\cite{seqtrack} that introduces temporal information by inputting more templates.
\begin{table}[t]
    \centering
        \fontsize{9}{11}\selectfont
        \begin{tabular}{c|ccccc}
        \toprule
         Model & Params& FLOPs  & Speed  \\
         \midrule
         SeqTrack-B256\cite{seqtrack} & 89M & 65G  & 40fps  \\
         SeqTrack-B384\cite{seqtrack} & 89M & 148G  & 15fps \\ 
         \midrule
         {\mytracker}-256(ours)   & 70M  & 24.8G & 46fps \\
         {\mytracker}-384(ours)   & 70M  & 55.7G & 36fps \\
        \bottomrule 
        \end{tabular}
    \caption{ Comparison of model Params, FLOPs, and Speed on NVIDIA V100.}
    \label{tab:fps}
\end{table} 

\subsection{Results and Comparisons.}
    
\begin{table*}[t]
\centering
\begin{adjustbox}{valign=c,max width=\textwidth}
    \fontsize{10}{11}\selectfont
    \begin{tabular}{r|c|ccc|ccc|ccc|ccc}
    \toprule
    \multicolumn{1}{c|}{\multirow{2}{*}{Method} }
    & \multicolumn{1}{c|}{\multirow{2}{*}{Source}} 
    & \multicolumn{3}{c|}{LaSOT} 
    & \multicolumn{3}{c|}{LaSOT$_{ext}$} 
    & \multicolumn{3}{c|}{GOT-10k$^*$} 
    & \multicolumn{3}{c}{TrackingNet}\\
    \cline{3-14}
                                            && AUC & P$_{norm}$ & P     & AUC & P$_{norm}$ & P      & AO & SR$_{0.5}$ & SR$_{0.75}$     & AUC & P$_{norm}$ & P \\
    \midrule

    \textbf{{\mytracker}-256} & Ours                 & \textbf{72.0} & \textbf{82.1} & \textbf{79.1}      &\textbf{52.4} & \textbf{63.3} & \textbf{60.2} & \textbf{74.9} & \textbf{84.8} & 71.7  & \textbf{84.3} & \textbf{88.8} & \textbf{83.5}\\
    \midrule
    AQATrack-256\cite{AQATrack} & CVPR24    & \underline{71.4} & \underline{81.9} & \underline{78.6}    & \underline{51.2} & \underline{62.2} & \underline{58.9}    & 73.8 & 83.2 & \textbf{72.1}    & 83.8 & 88.6 & \underline{83.1}\\
    EVPTrack-224\cite{evptrack} &AAAI24     & 70.4 & 80.9 & 77.2    & 48.7 & 59.5 & 55.1    & 73.3 & 83.6 & 70.7    & 83.5 & 88.3 & - \\
    F-BDMTrack-256\cite{F-BDMTrack}&ICCV23  & 69.9 & 79.4 & 75.8    & 47.9 & 57.9 & 54.0    & 72.7 & 82.0 & 69.9    & 83.7 & 88.3 & 82.6 \\
    ROMTrack-256\cite{ROMTrack} & ICCV23    & 69.3 & 78.8 & 75.6    & 48.9 & 59.3 & 55.0    & 72.9 & 82.9 & 70.2    & 83.6 & 88.4 & 82.7 \\
    ARTrack-256\cite{ARTrack} & CVPR23      & 70.4 & 79.5 & 76.6    & 46.4 & 56.5 & 52.3    & 73.5 & 82.2 & 70.9    & \underline{84.2} & \underline{88.7} & \textbf{83.5} \\
    SeqTrack-B256\cite{seqtrack} & CVPR23   & 69.9 & 79.7 & 76.3    & 49.5 & 60.8 & 56.3    & \underline{74.7} & \underline{84.7} & \underline{71.8}    & 83.3 & 88.3    & 82.2 \\
    VideoTrack\cite{VideoTrack1}  & CVPR23  & 70.2 & - & 76.4       & -    & -    & -       & 72.9 & 81.9 & 69.8    & 83.8 & \underline{88.7} & \underline{83.1} \\
    MixFormer-22k\cite{mixformer}&CVPR22    & 69.2 & 78.7 & 74.7    & -    & -    & -       & 70.7 & 80.0 & 67.8    & 83.1 & 88.1 & 81.6 \\
    OSTrack-256\cite{ostrack} &ECCV22       & 69.1 & 78.7 & 75.2    & 47.4 & 57.3 & 53.3    & 71.0 & 80.4 & 68.2    & 83.1 & 87.8 & 82.0\\
    STARK-ST101\cite{stark} & ICCV21        & 67.1 & 77.0 & -       & -    &-     & -       & 68.8 & 78.1 & 64.1    & 82.0 & 86.9 & -    \\
    TransT \cite{transt}& CVPR21            & 64.9 & 73.8 & 69.0    & -    & -    & -       & 67.1 & 76.8 & 60.9    & 81.4 & 86.7 & 80.3 \\
    Ocean \cite{Ocean}&  ECCV 20            & 56.0 & 65.1 & 56.6    &-     & -    &-        & 61.1 & 72.1 & 47.3    & -    & -    &-     \\
    SiamRPN++\cite{SiamRPN++}&CVPR19        & 49.6 & 56.9 & 49.1    & 34.0 & 41.6 & 39.6    & 51.7 & 61.6 & 32.5    & 73.3 & 80.0 & 69.4 \\
    ECO \cite{ECO} & ICCV 17                & 32.4 & 33.8 & 30.1    & 22.0 & 25.2 & 24.0    & 31.6 & 30.9 & 11.1    & -    & -    & - \\
    SiamFC \cite{SiamFC} & ECCVW16          & 33.6 & 42.0 & 33.9    & 23.0 & 31.1 & 26.9    & 34.8 & 35.3 & 9.8     & -    & -    & - \\
    \midrule
    
    \multicolumn{14}{l}{\multirow{1}{*}{\textit{Some Trackers with Higher Resolution}} }\\
    \midrule
    OSTrack-384\cite{ostrack}&ECCV22        & 71.1 & 81.1 & 77.6    & 50.5 & 61.3 & 57.6    & 73.7 & 83.2 & 70.8    & 83.9 & 88.5 & 83.2  \\
    ROMTrack-384\cite{ROMTrack} & ICCV23    & 71.4 & 81.4 & 78.2    & 51.3 & 62.4 & 58.6    & 74.2 & 84.3 & 72.4    & 84.1 & 89.0 & 83.7 \\
    F-BDMTrack-384\cite{F-BDMTrack}&ICCV23  & 72.0 & 81.5 & 77.7    & 50.8 & 61.3 & 57.8    & 75.4 & 84.3 & 72.9    & 84.5 & 89.0 & 84.0\\
    SeqTrack-B384\cite{seqtrack} & CVPR23   & 71.5 & 81.1 & 77.8    & 50.5 & 61.6 & 57.5    & 74.5 & 84.3 & 71.4    & 83.9 & 88.8 & 83.6 \\
    ARTrack-384\cite{ARTrack} & CVPR23      & 72.6 & 81.7 & 79.1    & 51.9 & 62.0 & 58.5    & 75.5 & 84.3 & 74.3    & \textbf{85.1} & \underline{89.1} & \textbf{84.8} \\
    HIPTrack\cite{hiptrack}    & CVPR24     & \underline{72.7} & \underline{82.9} & 79.5    & \underline{53.0} & \underline{64.3} & 60.6    & \textbf{77.4} & \textbf{88.0} & \underline{74.5}    & 84.5 & \underline{89.1} & 83.8\\
    AQATrack-384\cite{AQATrack} & CVPR24    & \underline{72.7} & \underline{82.9} & \underline{80.2}    & 52.7 & 64.2 & \underline{60.8}    & 76.0 & \underline{85.2} & \textbf{74.9}  & 84.8 & \textbf{89.3} & \underline{84.3}\\
    \midrule
    \textbf{{\mytracker}-384} & Ours        & \textbf{73.1} & \textbf{83.0} & \textbf{80.7}    & \textbf{53.4} & \textbf{64.8} & \textbf{61.0}    & \underline{76.1} & 84.9 & 74.4 & \underline{85.0} & \textbf{89.3} & \textbf{84.8}\\
    \bottomrule    
    \end{tabular}
    \end{adjustbox}
\caption{ Performance comparisons with state-of-the-art trackers on the test set of LaSOT\cite{lasot}, $\rm LaSOT_{ext}$\cite{lasot-ext} , GOT-10k\cite{got10k} and Trackingnet\cite{trackingnet}. We add a symbol * over GOT-10k to indicate that the corresponding models are only trained with the GOT-10k training set. The top two results are highlighted using \textbf{bold} and \underline{underlined} fonts respectively.}
\label{tab:1}
\end{table*} 
\begin{figure}[t]
    \centering
    \includegraphics[width=8.4cm, height=7.0cm]{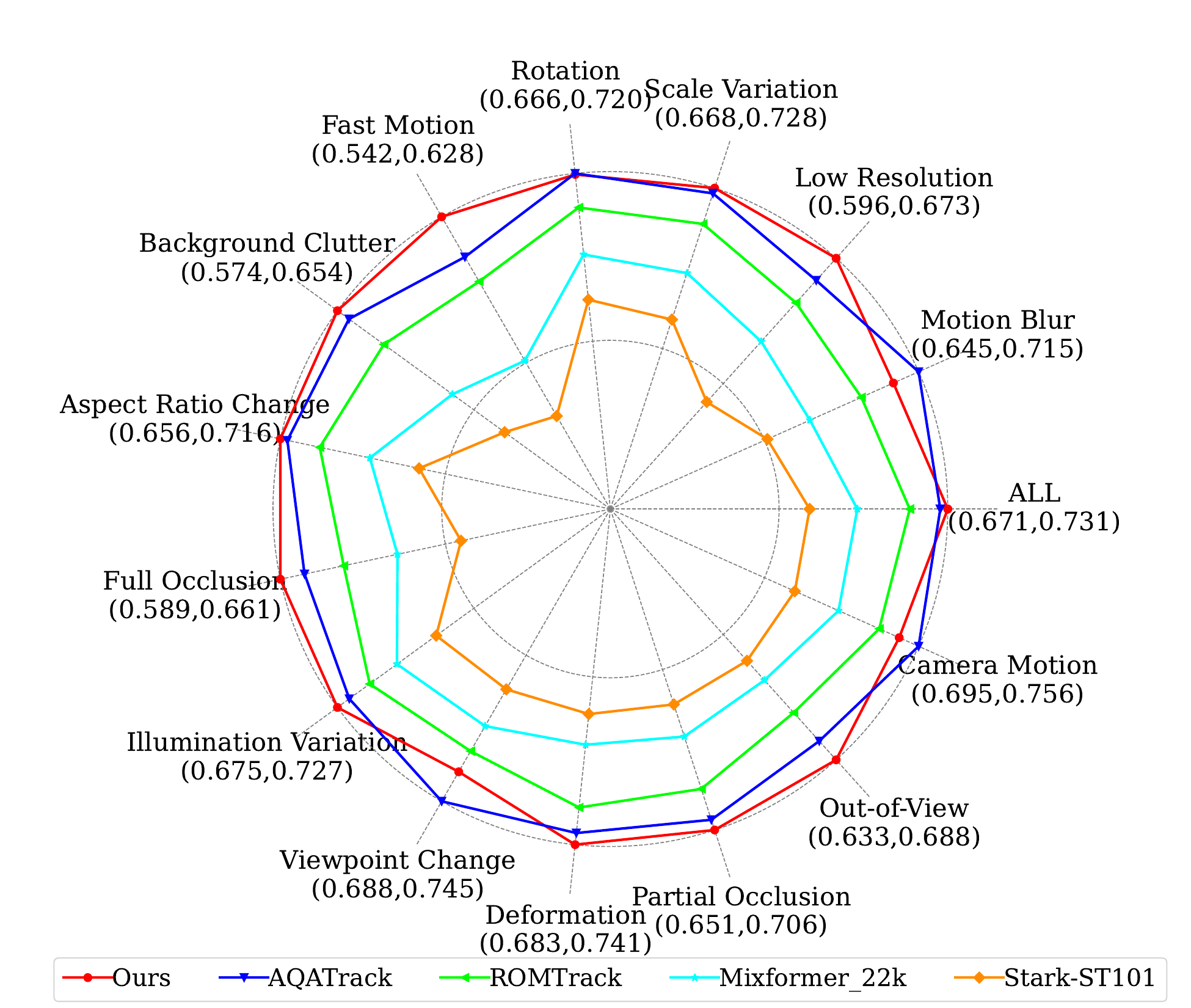}
    \caption{AUC scores of difference attributes on LaSOT\cite{lasot}. Best viewed in color.}
    \label{fig:zhizhuwang}
\end{figure}  
\begin{figure}[t]
    \centering
    \includegraphics[width=8.4cm,height=4.3cm]{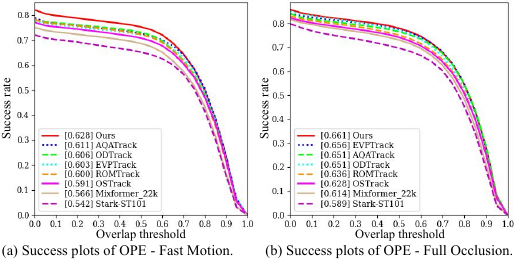}
    \caption{Success plots of one-pass evaluation (OPE) about fast motion and full occlusion challenges on LaSOT.
    Best viewed in color and zooming in.
    }
    \label{fig:quxian}
\end{figure} 
\begin{table*}[t]
    \centering
    \begin{adjustbox}{valign=c,max width=\textwidth}
        \fontsize{10}{11}\selectfont
        \begin{tabular}{c|cccccccccc|c}
        \toprule
                   &SiamFC &ECO   &SiamRPN++   &TransT &OSTrack & SeqTrack &ARTrack &F-BDMTrack &EVPTrack &AQATrack &{\mytracker} \\ 
        \cline{1-12} 
        UAV123       & 46.8  & 53.5 & 61.0     & 69.1  & 68.3   & 69.2     & 67.7   & 69.0      & 70.2    & \underline{70.7} & \textbf{70.8}                  \\
        TNL2K        & 29.5  & 32.6 & 41.3     & 50.7  & 54.3   & 54.9     & 57.5   & 56.4      & 57.5    & \underline{57.8}        & \textbf{58.8}                \\ 
        \bottomrule  
        \end{tabular}
    \end{adjustbox}
\caption{ Performance comparisons with state-of-the-art trackers on the TNL2K\cite{tnl2k}. The top two results are highlighted with \textbf{blod} and \underline{underlined} fonts respectively.}
\label{tab:2}
\end{table*} 
    We compare our evaluation results with other SOTA methods on six benchmarks to prove our effectiveness.
    \subsubsection{LaSOT\cite{lasot}.}
    LaSOT \cite{lasot} is a high-quality benchmark for long-term challenge on single object tracking.
    It consists of 1120 sequences for training and 280 sequences for testing.
    To show the robustness of our tracker, we compare our tracker with many SOTA trackers in \cref{fig1}.
    Benefiting from the {\mytoken} and {\mymodule}, {\mytracker} learn the appearance changes and motion trends well.
    {\mytracker} achieves a new state-of-art result. As shown in \cref{tab:1}, {\mytracker}-256 obtain 72.0\% of AUC, which outperforms AQATrack by 0.6\%.
    We compare {\mytracker}-384 with four famous trackers in different challenges of LaSOT in \cref{fig:zhizhuwang}.
    {\mytracker} outperforms others in many challenges, such as fast motion, low resolution, and full occlusion.
    As illuminated in \cref{fig:quxian}, {\mytracker} significantly outperforms other trackers when encountering fast motion and full occlusion, outperforming ODTrack by 2.2\% and 1.0\% of success rate.
    The above outstanding performances on this long-term benchmark show the effectiveness of {\mytracker} in temporal information learning.

\subsubsection{LaSOT$_{ext}$\cite{lasot-ext}.}
This benchmark is an expansion of LaSOT\cite{lasot} with additional 150 long-tem sequences, introducing many challenges, such as fast-moving small objects.
In \cref{tab:1}, we show the result of {\mytracker} that indicate our {\mytracker} outperform other trackers by a substantial margin, obtaining the highest AUC, P$_{nrom}$, and P.
{\mytracker} achieves 52.4\% of AUC, outperforming 1.2\% than AQATrack\cite{AQATrack}.
The excellent performances show our tracker not only mines temporal information but also addresses fast-moving small objects well.

\subsubsection{TrackingNet\cite{trackingnet}.}
TrackingNet is a large-scale tracking dataset with more than 30,000 sequences for training and 511 sequences for testing.
This benchmark focuses on some challenges when tracking objects in the wild, such as background clutter, full occlusion, and low resolution.
We show the result of {\mytracker} and some SOTA trackers on TrackingNet\cite{trackingnet} in \cref{tab:2}.
Our tracker achieves the 85.0\% of AUC score which demonstrates the robustness of {\mytracker} in the field. 

\subsubsection{GOT-10k\cite{got10k}.}
GOT-10k is a large high-diversity benchmark for generic object tracking, which introduces a one-shot protocol for evaluation, i.e., the training and test classes are zero-overlapped.
Adhering to this protocol to train our tracker, we evaluate the tracker on GOT-10k to demonstrate our generalization.
As shown in \cref{tab:1}, our {\mytracker} achieves a competitive performance among state-of-art trackers.
The high performance on this one-shot tracking benchmark demonstrates the strong discriminative ability of {\mytracker} for unseen classes.

\subsubsection{UAV123\cite{uav123} and TNL2K\cite{tnl2k}.}
We also evaluate our tracker on two additional benchmarks: UAV123 and TNL2K.
They include 123 and 700 videos for testing, respectively.
As shown in \cref{tab:2}, our {\mytracker} with the lower resolution of search image achieves 70.8\% of AUC on UAV123 and 58.8\% of AUC on TNL2K, which are better than others.

\subsection{Ablation Study and Analysis.}\label{subsec:ablation}
\begin{table}[t]
    \centering
    \fontsize{9}{11}\selectfont
    \begin{tabular}{c|cc|cc}
    \toprule 
     \multirow{2}{*}{Method}  &\multicolumn{2}{c|}{LaSOT} &\multicolumn{2}{c}{GOT-10k}  \\
     \cline{2-5}
     & AUC & P$_{norm}$ & AO & SR$_{0.5}$  \\
     \midrule
     Baseline  & 71.1  & 81.2 & 73.0 & 82.8 \\
     +{\mytoken} & 71.4 & 81.5 & 73.7 & 83.3 \\ 
     +{\mymodule} & \textbf{72.0} & \textbf{82.1} & \textbf{74.9} & \textbf{84.8}  \\
    \bottomrule 
    \end{tabular}
    \caption{Ablation studies of {\mytracker} on different dataset.}
    \label{tab:ablation}
\end{table} 
\begin{table}[]
    \centering
    \fontsize{10}{9}\selectfont
    \begin{tabularx}{\linewidth}{>{\centering\arraybackslash}p{2cm}|>{\centering\arraybackslash}X>{\centering\arraybackslash}X>{\centering\arraybackslash}X}
    \toprule
         Component   & AO & SR$_{0.5}$ & SR$_{0.75}$ \\
         \midrule
         Baseline              & 73.0 & 82.8 & 71.6 \\
         Mamba\_Cross          & \textbf{74.9} & \textbf{84.8} & \textbf{71.7} \\
         Self\_Cross           & 74.3 & 84.6 & 71.7\\
         Self\_Self            & 74.2 & 84.2 & 70.8\\
    \bottomrule 
    \end{tabularx}
    \caption{Influence of different layers on GOT-10k.}
    \label{tab:temporal_module}
\end{table}
\begin{table}
    \centering
        \fontsize{9}{9}\selectfont
        \begin{tabularx}{\linewidth}{p{1.3cm}|>{\centering\arraybackslash}p{1.65cm}|>{\centering\arraybackslash}X>{\centering\arraybackslash}X>{\centering\arraybackslash}X}
        \toprule
         Backbone &Our\_method  & AUC  & P$_{norm}$ & P\\
         \midrule
         \multirow{2}{*}{ViT-B} & - & 68.6  & 78.4 & 74.3\\
                                & \checkmark & 69.6(+1.0)  & 79.7(+1.3) &75.5(+1.2)\\
         \midrule
         \multirow{2}{*}{HiViT} & - & 70.2 & 80.3 & 76.9 \\ 
         & \checkmark & 70.8(+0.6) & 80.9(+0.6) & 77.8(+0.9)\\
         \midrule
         \multirow{2}{*}{Fast-iTPN}  & -  & 71.1 & 81.2 & 78.2\\
         & \checkmark & \textbf{72.0}(+0.9) & \textbf{82.1}(+0.9) & \textbf{79.1}(+0.9) \\
        \bottomrule 
        \end{tabularx}
\caption{Influence of the backbone on LaSOT.}
\label{tab:backbone}
\end{table}
To demonstrate the effectiveness of our proposed method, we design ablation experiments from four aspects, namely ablation of {\mytracker}, different backbone, different components of {\mymodule}, and the size of the sliding window.
All the ablation study is based {\mytracker}-256.

\textbf{Ablation Studies of {\mytracker}.}
We explore the impact of each component used in {\mytracker} on LaSOT\cite{lasot} and GOT-10k\cite{got10k}, as shown in \cref{tab:ablation}.
The baseline based Fast-iTPN\cite{fastitpn} consists of a backbone and a head network.
For the sake of fairness, we keep the same config as the baseline for the following experiments.
Firstly, we show the impact of tokens' guidance in the absence of temporal information, which outperforms 0.3\% of AUC on LaSOT and outperforms 0.7\% of AO on GOT-10k  than the baseline.
These results show that the {\mytoken} can learn the target's appearance during interaction in the backbone, and help improve the expressive ability of search features.
Then, we introduce the temporal information extracted by the {\mymodule}.
The results show that temporal information improves the model's discriminative ability, which achieves 72.0\% of AUC on LaSOT and outperforms 1.9\% on GOT-10k\cite{got10k}.

\textbf{Variants of {\mymodule}.}
Our {\mymodule} consists of two layers, the first layer is mamba, and the second layer is cross-attention.
To demonstrate the effectiveness of the proposed temporal information learning method, we conduct experiments using different technical approaches.
Firstly, we demonstrate that using self-attention can also achieve good performance, achieving $74.3\%$ of AO in GOT-10k, which outperforms most trackers, as demonstrated in \cref{tab:temporal_module}.
Additionally, we demonstrate the performance when both two layers are implemented using self-attention.
The results are shown in the last row of \cref{tab:temporal_module}, achieving $74.2\%$ of AO in GOT-10k.
Although these variants all get a comparative result, the variant that introduces the mamba with autoregressive characteristic achieves the best performance.
\begin{figure}[t] 
    \centering
    \includegraphics[height=5.2cm]{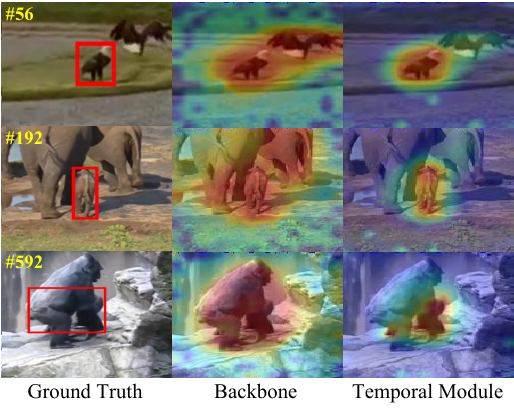}
    \caption{
    Visualize the attention of search to {\mytoken}. 
    The first column is ground truth, the second column is the attention in the last layer of the backbone, and the third column is the attention in {\mymodule}.
    }
    \label{fig-attn}
\end{figure}
\textbf{Different Sizes of the Sliding Window.}
The size of the window indicates the length of temporal information.
To explore the model's potential for mining temporal information, we design different sliding window sizes $m$, as illuminated in \cref{tab:temporal_number}. 
When the window size is $2$, the model learns a short temporal information, leading to a lower performance.
When we set $m$ to $4$, the model achieves $71.9\%$ of AUC on LaSOT\cite{lasot}.
When we set $m$ to 8, the model achieves $72.0\%$ of AUC.
Therefore, the optimal window size may be between $4$ and $8$.

\textbf{Different Backbone.}
We prove the effect of our method by replacing different backbones, such as ViT\cite{vit} in many trackers\cite{ostrack, simtrack} and HiViT\cite{hivit} used in some recent trackers\cite{evptrack, AQATrack}.
As shown in \cref{tab:backbone}, our method based on ViT achieves $69.6\%$ of AUC, improving by $1.0\%$.
Otherwise, our method used HiViT as backbone achieves $70.8\%$ of AUC, improving by $0.6\%$.
Our method based on Fast-iTPN improves the AUC by $0.9\%$.
The above results show the effectiveness of our method.

\textbf{Visualization and Qualitative Comparison.}
\begin{table}[]
    \centering
        \fontsize{10}{9}\selectfont
        \begin{tabularx}{0.45\textwidth}{>{\centering\arraybackslash}X|>{\centering\arraybackslash}X|>{\centering\arraybackslash}X>{\centering\arraybackslash}X>{\centering\arraybackslash}X} 
        \toprule
         m & n & AUC & P$_{norm}$  & P \\
         \midrule
         2  & 16 & 71.1 & 81.2 & 78.2\\
         4  & 8  & 71.9 & 82.0 & 79.0\\
         8  & 4  & \textbf{72.0}  & \textbf{82.1} & \textbf{79.1}\\
        \bottomrule 
        \end{tabularx}
    \caption{Influence of window size on LaSOT.}
    \label{tab:temporal_number}
\end{table}
Due to the backbone focusing more on appearance modeling and introducing temporal information, {\mytracker} achieves the most accurate tracking in the above challenging scenes.
We visualize the attention of search to the {\mytoken} in backbone and {\mymodule}, as demonstrated in \cref{fig-attn}.
In the third column, the search feature after guiding by {\mytoken} indicates a more accurate location of the target in similar object interference (first row) and occlusion (last row) cases.
\section{Conclusion}
We propose a novel tracker that elegantly extracts temporal information from a list of {\mytoken}s rather than several images, reducing the model's learning and computational burden.
The model's backbone focuses more on appearance modeling. 
Under the guidance of {\mytoken} contained temporal information, the appearance features adjust to obtain more accurate tracking results.
Extensive experiments on six datasets demonstrate the superiority of our method.

\section{Acknowledgements}
This work is supported by the National Natural Science Foundation of China (No.U23A20383, 62472109, and 62466051), the Project of Guangxi Science and Technology (No.2024GXNSFGA010001 and 2022GXNSFDA035079), the Guangxi ``Young Bagui Scholar” Teams for Innovation and Research Project, the Research Project of Guangxi Normal University (No.2024DF001).

\bibliography{aaai25.bib}

\begin{thebibliography}{61}
\providecommand{\natexlab}[1]{#1}

\bibitem[{Bai et~al.(2024)Bai, Zhao, Gong, and Wei}]{artrackv2}
Bai, Y.; Zhao, Z.; Gong, Y.; and Wei, X. 2024.
\newblock ARTrackV2: Prompting Autoregressive Tracker Where to Look and How to Describe.
\newblock In \emph{Proceedings of the IEEE/CVF Conference on Computer Vision and Pattern Recognition (CVPR)}.

\bibitem[{Bertinetto et~al.(2016)Bertinetto, Valmadre, Henriques, Vedaldi, and Torr}]{SiamFC}
Bertinetto, L.; Valmadre, J.; Henriques, J.~F.; Vedaldi, A.; and Torr, P. H.~S. 2016.
\newblock Fully-Convolutional Siamese Networks for Object Tracking.
\newblock In \emph{{ECCV} Workshops}, 850--865.

\bibitem[{Cai, Liu, and Wang(2024)}]{hiptrack}
Cai, W.; Liu, Q.; and Wang, Y. 2024.
\newblock HIPTrack: Visual Tracking with Historical Prompts.
\newblock In \emph{Proceedings of the IEEE/CVF Conference on Computer Vision and Pattern Recognition}.

\bibitem[{Cai et~al.(2023)Cai, Liu, Tang, and Wu}]{ROMTrack}
Cai, Y.; Liu, J.; Tang, J.; and Wu, G. 2023.
\newblock Robust Object Modeling for Visual Tracking.
\newblock \emph{CoRR}, abs/2308.05140.

\bibitem[{Cao et~al.(2023)Cao, Huang, Pan, Zhang, Liu, and Fu}]{TCTrack++}
Cao, Z.; Huang, Z.; Pan, L.; Zhang, S.; Liu, Z.; and Fu, C. 2023.
\newblock Towards real-world visual tracking with temporal contexts.
\newblock \emph{IEEE Transactions on Pattern Analysis and Machine Intelligence}.

\bibitem[{Chen et~al.(2022)Chen, Li, Bai, Qiao, Shen, Li, Gan, Wu, and Ouyang}]{simtrack}
Chen, B.; Li, P.; Bai, L.; Qiao, L.; Shen, Q.; Li, B.; Gan, W.; Wu, W.; and Ouyang, W. 2022.
\newblock Backbone is All Your Need: {A} Simplified Architecture for Visual Object Tracking.
\newblock In \emph{{ECCV} {(22)}}, 375--392.

\bibitem[{Chen et~al.(2023)Chen, Peng, Wang, Lu, and Hu}]{seqtrack}
Chen, X.; Peng, H.; Wang, D.; Lu, H.; and Hu, H. 2023.
\newblock Seqtrack: Sequence to sequence learning for visual object tracking.
\newblock In \emph{Proceedings of the IEEE/CVF conference on computer vision and pattern recognition}, 14572--14581.

\bibitem[{Chen et~al.(2021)Chen, Yan, Zhu, Wang, Yang, and Lu}]{transt}
Chen, X.; Yan, B.; Zhu, J.; Wang, D.; Yang, X.; and Lu, H. 2021.
\newblock Transformer Tracking.
\newblock In \emph{{CVPR}}, 8126--8135.

\bibitem[{Chen et~al.(2020)Chen, Zhong, Li, Zhang, and Ji}]{Siamban}
Chen, Z.; Zhong, B.; Li, G.; Zhang, S.; and Ji, R. 2020.
\newblock Siamese Box Adaptive Network for Visual Tracking.
\newblock In \emph{{CVPR}}, 6667--6676.

\bibitem[{Cheng, Wang, and Li(2022)}]{surveillance0}
Cheng, L.; Wang, J.; and Li, Y. 2022.
\newblock ViTrack: Efficient Tracking on the Edge for Commodity Video Surveillance Systems.
\newblock \emph{IEEE Transactions on Parallel and Distributed Systems}, 33(3): 723--735.

\bibitem[{Cui et~al.(2022)Cui, Jiang, Wang, and Wu}]{mixformer}
Cui, Y.; Jiang, C.; Wang, L.; and Wu, G. 2022.
\newblock MixFormer: End-to-End Tracking with Iterative Mixed Attention.
\newblock In \emph{{CVPR}}, 13598--13608.

\bibitem[{Cui et~al.(2024)Cui, Jiang, Wu, and Wang}]{MixViT}
Cui, Y.; Jiang, C.; Wu, G.; and Wang, L. 2024.
\newblock MixFormer: End-to-End Tracking With Iterative Mixed Attention.
\newblock \emph{IEEE Transactions on Pattern Analysis and Machine Intelligence}, 46(6): 4129--4146.

\bibitem[{Danelljan et~al.(2017)Danelljan, Bhat, Khan, and Felsberg}]{ECO}
Danelljan, M.; Bhat, G.; Khan, F.~S.; and Felsberg, M. 2017.
\newblock {ECO:} Efficient Convolution Operators for Tracking.
\newblock In \emph{{CVPR}}, 6931--6939.

\bibitem[{Dosovitskiy et~al.(2021)Dosovitskiy, Beyer, Kolesnikov, Weissenborn, Zhai, Unterthiner, Dehghani, Minderer, Heigold, Gelly, Uszkoreit, and Houlsby}]{vit}
Dosovitskiy, A.; Beyer, L.; Kolesnikov, A.; Weissenborn, D.; Zhai, X.; Unterthiner, T.; Dehghani, M.; Minderer, M.; Heigold, G.; Gelly, S.; Uszkoreit, J.; and Houlsby, N. 2021.
\newblock An Image is Worth 16x16 Words: Transformers for Image Recognition at Scale.
\newblock In \emph{{ICLR}}.

\bibitem[{Fan et~al.(2021)Fan, Bai, Lin, Yang, Chu, Deng, Yu, Harshit, Huang, Liu, Xu, Liao, Yuan, and Ling}]{lasot-ext}
Fan, H.; Bai, H.; Lin, L.; Yang, F.; Chu, P.; Deng, G.; Yu, S.; Harshit; Huang, M.; Liu, J.; Xu, Y.; Liao, C.; Yuan, L.; and Ling, H. 2021.
\newblock LaSOT: {A} High-quality Large-scale Single Object Tracking Benchmark.
\newblock \emph{Int. J. Comput. Vis.}, 439--461.

\bibitem[{Fan et~al.(2019)Fan, Lin, Yang, Chu, Deng, Yu, Bai, Xu, Liao, and Ling}]{lasot}
Fan, H.; Lin, L.; Yang, F.; Chu, P.; Deng, G.; Yu, S.; Bai, H.; Xu, Y.; Liao, C.; and Ling, H. 2019.
\newblock LaSOT: {A} High-Quality Benchmark for Large-Scale Single Object Tracking.
\newblock In \emph{{CVPR}}, 5374--5383.

\bibitem[{Fu et~al.(2022)Fu, Fu, Liu, Cai, and Wang}]{sparseTT}
Fu, Z.; Fu, Z.; Liu, Q.; Cai, W.; and Wang, Y. 2022.
\newblock Sparsett: Visual tracking with sparse transformers.
\newblock \emph{arXiv preprint arXiv:2205.03776}.

\bibitem[{Fu et~al.(2021)Fu, Liu, Fu, and Wang}]{STMTrack}
Fu, Z.; Liu, Q.; Fu, Z.; and Wang, Y. 2021.
\newblock STMTrack: Template-Free Visual Tracking With Space-Time Memory Networks.
\newblock In \emph{{CVPR}}, 13774--13783.

\bibitem[{Gao et~al.(2022)Gao, Zhou, Ma, Wang, and Yuan}]{aiatrack}
Gao, S.; Zhou, C.; Ma, C.; Wang, X.; and Yuan, J. 2022.
\newblock AiATrack: Attention in Attention for Transformer Visual Tracking.
\newblock In \emph{{ECCV} {(22)}}, 146--164.

\bibitem[{Gao, Zhou, and Zhang(2023)}]{GRM}
Gao, S.; Zhou, C.; and Zhang, J. 2023.
\newblock Generalized Relation Modeling for Transformer Tracking.
\newblock In \emph{Proceedings of the IEEE/CVF Conference on Computer Vision and Pattern Recognition}, 18686--18695.

\bibitem[{Gu and Dao(2023)}]{mamba}
Gu, A.; and Dao, T. 2023.
\newblock Mamba: Linear-time sequence modeling with selective state spaces.
\newblock \emph{arXiv preprint arXiv:2312.00752}.

\bibitem[{He et~al.(2016)He, Zhang, Ren, and Sun}]{resnet1}
He, K.; Zhang, X.; Ren, S.; and Sun, J. 2016.
\newblock Deep Residual Learning for Image Recognition.
\newblock In \emph{Proceedings of the IEEE Conference on Computer Vision and Pattern Recognition (CVPR)}.

\bibitem[{Hu et~al.(2024{\natexlab{a}})Hu, Zhong, Liang, Zhang, Li, and Li}]{hu2024}
Hu, X.; Zhong, B.; Liang, Q.; Zhang, S.; Li, N.; and Li, X. 2024{\natexlab{a}}.
\newblock Towards Modalities Correlation for RGB-T Tracking.
\newblock \emph{IEEE Transactions on Circuits and Systems for Video Technology}.

\bibitem[{Hu et~al.(2024{\natexlab{b}})Hu, Zhong, Liang, Zhang, Li, Li, and Ji}]{huxiantao}
Hu, X.; Zhong, B.; Liang, Q.; Zhang, S.; Li, N.; Li, X.; and Ji, R. 2024{\natexlab{b}}.
\newblock Transformer Tracking via Frequency Fusion.
\newblock \emph{IEEE Transactions on Circuits and Systems for Video Technology}, 34(2): 1020--1031.

\bibitem[{Huang, Zhao, and Huang(2021)}]{got10k}
Huang, L.; Zhao, X.; and Huang, K. 2021.
\newblock GOT-10k: {A} Large High-Diversity Benchmark for Generic Object Tracking in the Wild.
\newblock \emph{{IEEE} Trans. Pattern Anal. Mach. Intell.}, 43(5): 1562--1577.

\bibitem[{Huang et~al.(2024)Huang, Pei, You, Wang, Qian, and Xu}]{localmamba}
Huang, T.; Pei, X.; You, S.; Wang, F.; Qian, C.; and Xu, C. 2024.
\newblock Localmamba: Visual state space model with windowed selective scan.
\newblock \emph{arXiv preprint arXiv:2403.09338}.

\bibitem[{Krizhevsky, Sutskever, and Hinton(2012)}]{alexnet}
Krizhevsky, A.; Sutskever, I.; and Hinton, G.~E. 2012.
\newblock ImageNet Classification with Deep Convolutional Neural Networks.
\newblock In \emph{{NIPS}}, 1106--1114.

\bibitem[{Li et~al.(2019)Li, Wu, Wang, Zhang, Xing, and Yan}]{SiamRPN++}
Li, B.; Wu, W.; Wang, Q.; Zhang, F.; Xing, J.; and Yan, J. 2019.
\newblock SiamRPN++: Evolution of Siamese Visual Tracking With Very Deep Networks.
\newblock In \emph{{CVPR}}, 4282--4291.

\bibitem[{Lin et~al.(2022)Lin, Fan, Zhang, Xu, and Ling}]{SwinTrack}
Lin, L.; Fan, H.; Zhang, Z.; Xu, Y.; and Ling, H. 2022.
\newblock Swintrack: A simple and strong baseline for transformer tracking.
\newblock \emph{Advances in Neural Information Processing Systems}, 35: 16743--16754.

\bibitem[{Lin et~al.(2017)Lin, Goyal, Girshick, He, and Doll{\'{a}}r}]{focalloss}
Lin, T.; Goyal, P.; Girshick, R.~B.; He, K.; and Doll{\'{a}}r, P. 2017.
\newblock Focal Loss for Dense Object Detection.
\newblock In \emph{{ICCV}}, 2999--3007.

\bibitem[{Lin et~al.(2014)Lin, Maire, Belongie, Hays, Perona, Ramanan, Doll{\'{a}}r, and Zitnick}]{coco}
Lin, T.; Maire, M.; Belongie, S.~J.; Hays, J.; Perona, P.; Ramanan, D.; Doll{\'{a}}r, P.; and Zitnick, C.~L. 2014.
\newblock Microsoft {COCO:} Common Objects in Context.
\newblock In \emph{{ECCV}}, 740--755.

\bibitem[{Liu et~al.(2024)Liu, Tian, Zhao, Yu, Xie, Wang, Ye, and Liu}]{vmamba}
Liu, Y.; Tian, Y.; Zhao, Y.; Yu, H.; Xie, L.; Wang, Y.; Ye, Q.; and Liu, Y. 2024.
\newblock VMamba: Visual State Space Model.
\newblock \emph{arXiv preprint arXiv:2401.10166}.

\bibitem[{Liu et~al.(2021)Liu, Lin, Cao, Hu, Wei, Zhang, Lin, and Guo}]{swintransformer}
Liu, Z.; Lin, Y.; Cao, Y.; Hu, H.; Wei, Y.; Zhang, Z.; Lin, S.; and Guo, B. 2021.
\newblock Swin Transformer: Hierarchical Vision Transformer using Shifted Windows.
\newblock In \emph{{ICCV}}, 9992--10002. {IEEE}.

\bibitem[{Loshchilov and Hutter(2019)}]{adamw}
Loshchilov, I.; and Hutter, F. 2019.
\newblock Decoupled Weight Decay Regularization.
\newblock In \emph{{ICLR}}.

\bibitem[{Ma, Li, and Wang(2024)}]{u-mamba}
Ma, J.; Li, F.; and Wang, B. 2024.
\newblock U-mamba: Enhancing long-range dependency for biomedical image segmentation.
\newblock \emph{arXiv preprint arXiv:2401.04722}.

\bibitem[{Mueller, Smith, and Ghanem(2016)}]{uav123}
Mueller, M.; Smith, N.; and Ghanem, B. 2016.
\newblock A Benchmark and Simulator for {UAV} Tracking.
\newblock In \emph{{ECCV}}, 445--461.

\bibitem[{M{\"{u}}ller et~al.(2018)M{\"{u}}ller, Bibi, Giancola, Al{-}Subaihi, and Ghanem}]{trackingnet}
M{\"{u}}ller, M.; Bibi, A.; Giancola, S.; Al{-}Subaihi, S.; and Ghanem, B. 2018.
\newblock TrackingNet: {A} Large-Scale Dataset and Benchmark for Object Tracking in the Wild.
\newblock In \emph{{ECCV}}, 310--327.

\bibitem[{Pereira et~al.(2022)Pereira, Carvalho, Garrote, and Nunes}]{mobile_robotics}
Pereira, R.; Carvalho, G.; Garrote, L.; and Nunes, U.~J. 2022.
\newblock Sort and deep-SORT based multi-object tracking for mobile robotics: Evaluation with new data association metrics.
\newblock \emph{Applied Sciences}, 12(3): 1319.

\bibitem[{Premachandra, Ueda, and Suzuki(2020)}]{autodriving0}
Premachandra, C.; Ueda, S.; and Suzuki, Y. 2020.
\newblock Detection and Tracking of Moving Objects at Road Intersections Using a 360-Degree Camera for Driver Assistance and Automated Driving.
\newblock \emph{IEEE Access}, 8: 135652--135660.

\bibitem[{Rezatofighi et~al.(2019)Rezatofighi, Tsoi, Gwak, Sadeghian, Reid, and Savarese}]{giou}
Rezatofighi, H.; Tsoi, N.; Gwak, J.; Sadeghian, A.; Reid, I.~D.; and Savarese, S. 2019.
\newblock Generalized Intersection Over Union: {A} Metric and a Loss for Bounding Box Regression.
\newblock In \emph{{CVPR}}, 658--666.

\bibitem[{Shehzed, Jalal, and Kim(2019)}]{surveillance1}
Shehzed, A.; Jalal, A.; and Kim, K. 2019.
\newblock Multi-Person Tracking in Smart Surveillance System for Crowd Counting and Normal/Abnormal Events Detection.
\newblock In \emph{2019 International Conference on Applied and Engineering Mathematics (ICAEM)}, 163--168.

\bibitem[{Shi et~al.(2024)Shi, Zhong, Liang, Li, Zhang, and Li}]{evptrack}
Shi, L.; Zhong, B.; Liang, Q.; Li, N.; Zhang, S.; and Li, X. 2024.
\newblock Explicit Visual Prompts for Visual Object Tracking.
\newblock In \emph{Proceedings of the AAAI Conference on Artificial Intelligence}, volume~38, 4838--4846.

\bibitem[{Song et~al.(2023)Song, Luo, Yu, Chen, and Yang}]{CTTrack}
Song, Z.; Luo, R.; Yu, J.; Chen, Y.-P.~P.; and Yang, W. 2023.
\newblock Compact Transformer Tracker with Correlative Masked Modeling.
\newblock In \emph{Proceedings of the AAAI Conference on Artificial Intelligence (AAAI)}.

\bibitem[{Song et~al.(2022)Song, Yu, Chen, and Yang}]{cswintt}
Song, Z.; Yu, J.; Chen, Y.-P.~P.; and Yang, W. 2022.
\newblock Transformer tracking with cyclic shifting window attention.
\newblock In \emph{Proceedings of the IEEE/CVF conference on computer vision and pattern recognition}, 8791--8800.

\bibitem[{Tian et~al.(2024)Tian, Xie, Qiu, Jiao, Wang, Tian, and Ye}]{fastitpn}
Tian, Y.; Xie, L.; Qiu, J.; Jiao, J.; Wang, Y.; Tian, Q.; and Ye, Q. 2024.
\newblock Fast-iTPN: Integrally Pre-Trained Transformer Pyramid Network with Token Migration.
\newblock arXiv:2211.12735.

\bibitem[{Wang et~al.(2021)Wang, Shu, Zhang, Jiang, Wang, Tian, and Wu}]{tnl2k}
Wang, X.; Shu, X.; Zhang, Z.; Jiang, B.; Wang, Y.; Tian, Y.; and Wu, F. 2021.
\newblock Towards More Flexible and Accurate Object Tracking With Natural Language: Algorithms and Benchmark.
\newblock In \emph{{CVPR}}, 13763--13773.

\bibitem[{Wei et~al.(2023)Wei, Bai, Zheng, Shi, and Gong}]{ARTrack}
Wei, X.; Bai, Y.; Zheng, Y.; Shi, D.; and Gong, Y. 2023.
\newblock Autoregressive visual tracking.
\newblock In \emph{Proceedings of the IEEE/CVF Conference on Computer Vision and Pattern Recognition}, 9697--9706.

\bibitem[{Xie et~al.(2023)Xie, Chu, Li, Lu, and Ma}]{VideoTrack1}
Xie, F.; Chu, L.; Li, J.; Lu, Y.; and Ma, C. 2023.
\newblock VideoTrack: Learning to Track Objects via Video Transformer.
\newblock In \emph{Proceedings of the IEEE/CVF Conference on Computer Vision and Pattern Recognition (CVPR)}, 22826--22835.

\bibitem[{Xie et~al.(2024)Xie, Zhong, Mo, Zhang, Shi, Song, and Ji}]{AQATrack}
Xie, J.; Zhong, B.; Mo, Z.; Zhang, S.; Shi, L.; Song, S.; and Ji, R. 2024.
\newblock Autoregressive Queries for Adaptive Tracking with Spatio-Temporal Transformers.
\newblock In \emph{Proceedings of the IEEE/CVF Conference on Computer Vision and Pattern Recognition}, 19300--19309.

\bibitem[{Xing et~al.(2024)Xing, Ye, Yang, Liu, and Zhu}]{segmamba}
Xing, Z.; Ye, T.; Yang, Y.; Liu, G.; and Zhu, L. 2024.
\newblock Segmamba: Long-range sequential modeling mamba for 3d medical image segmentation.
\newblock \emph{arXiv preprint arXiv:2401.13560}.

\bibitem[{Xu et~al.(2020)Xu, Wang, Li, Ye, and Yu}]{SiamFC++}
Xu, Y.; Wang, Z.; Li, Z.; Ye, Y.; and Yu, G. 2020.
\newblock SiamFC++: Towards Robust and Accurate Visual Tracking with Target Estimation Guidelines.
\newblock In \emph{{AAAI}}, 12549--12556.

\bibitem[{Xue et~al.(2024)Xue, Zhong, Liang, Xia, and Song}]{xue2024}
Xue, C.; Zhong, B.; Liang, Q.; Xia, H.; and Song, S. 2024.
\newblock Unifying Motion and Appearance Cues for Visual Tracking via Shared Queries.
\newblock \emph{IEEE Transactions on Circuits and Systems for Video Technology}.

\bibitem[{Yan et~al.(2021)Yan, Peng, Fu, Wang, and Lu}]{stark}
Yan, B.; Peng, H.; Fu, J.; Wang, D.; and Lu, H. 2021.
\newblock Learning Spatio-Temporal Transformer for Visual Tracking.
\newblock In \emph{ICCV}, 10428--10437.

\bibitem[{Yang et~al.(2023)Yang, He, Ma, Yu, and Zhang}]{F-BDMTrack}
Yang, D.; He, J.; Ma, Y.; Yu, Q.; and Zhang, T. 2023.
\newblock Foreground-Background Distribution Modeling Transformer for Visual Object Tracking.
\newblock In \emph{Proceedings of the IEEE/CVF International Conference on Computer Vision (ICCV)}, 10117--10127.

\bibitem[{Ye et~al.(2022)Ye, Chang, Ma, Shan, and Chen}]{ostrack}
Ye, B.; Chang, H.; Ma, B.; Shan, S.; and Chen, X. 2022.
\newblock Joint Feature Learning and Relation Modeling for Tracking: {A} One-Stream Framework.
\newblock In \emph{{ECCV} {(22)}}, 341--357.

\bibitem[{Yu and Wang(2024)}]{mambaout}
Yu, W.; and Wang, X. 2024.
\newblock MambaOut: Do We Really Need Mamba for Vision?
\newblock \emph{arXiv preprint arXiv:2405.07992}.

\bibitem[{Zhang et~al.(2019)Zhang, Gonzalez{-}Garcia, van~de Weijer, Danelljan, and Khan}]{updateNet}
Zhang, L.; Gonzalez{-}Garcia, A.; van~de Weijer, J.; Danelljan, M.; and Khan, F.~S. 2019.
\newblock Learning the Model Update for Siamese Trackers.
\newblock In \emph{{ICCV}}, 4009--4018.

\bibitem[{Zhang et~al.(2023)Zhang, Tian, Xie, Huang, Dai, Ye, and Tian}]{hivit}
Zhang, X.; Tian, Y.; Xie, L.; Huang, W.; Dai, Q.; Ye, Q.; and Tian, Q. 2023.
\newblock HiViT: A Simpler and More Efficient Design of Hierarchical Vision Transformer.
\newblock In \emph{International Conference on Learning Representations}.

\bibitem[{Zhang et~al.(2020)Zhang, Peng, Fu, Li, and Hu}]{Ocean}
Zhang, Z.; Peng, H.; Fu, J.; Li, B.; and Hu, W. 2020.
\newblock Ocean: Object-Aware Anchor-Free Tracking.
\newblock In \emph{{ECCV}}, 771--787.

\bibitem[{Zheng et~al.(2024)Zheng, Zhong, Liang, Mo, Zhang, and Li}]{odtrack}
Zheng, Y.; Zhong, B.; Liang, Q.; Mo, Z.; Zhang, S.; and Li, X. 2024.
\newblock Odtrack: Online dense temporal token learning for visual tracking.
\newblock In \emph{Proceedings of the AAAI Conference on Artificial Intelligence}, volume~38, 7588--7596.

\bibitem[{Zhu et~al.(2024)Zhu, Liao, Zhang, Wang, Liu, and Wang}]{vim}
Zhu, L.; Liao, B.; Zhang, Q.; Wang, X.; Liu, W.; and Wang, X. 2024.
\newblock Vision mamba: Efficient visual representation learning with bidirectional state space model.
\newblock \emph{arXiv preprint arXiv:2401.09417}.

\end{thebibliography}

\end{document}